\documentclass[11pt,a4paper]{article}
\usepackage[hyperref]{acl2021}
\usepackage{times}
\usepackage{latexsym}
\usepackage{booktabs}
\usepackage{amsfonts}

\usepackage{microtype}
\usepackage{subfigure}
\usepackage{amsmath}
\usepackage{graphicx}

\aclfinalcopy 
\def\aclpaperid{168} 

\newcommand{\bertcls}{\mathrm{cls}}
\newcommand{\bertsep}{\mathrm{sep}}
\newcommand{\bh}{\mathbf{H}}
\newcommand{\mb}{\mathbf}

\newcommand{\mtd}{REPT}
\newcommand{\modb}{BERT-Q~}
\newcommand{\modrb}{RoBERTa-Q~}

\title{REPT: Bridging Language Models and Machine Reading Comprehension via Retrieval-Based Pre-training}

\author{Fangkai Jiao$^{1}$\footnotemark[1], Yangyang Guo$^1$, Yilin Niu$^2$, Feng Ji$^3$, Feng-Lin Li$^3$, Liqiang Nie$^1$\footnotemark[2] \\
  $^1$ School of Computer Science and Technology, Shandong University, Qingdao, China \\
  $^2$ Department of Computer Science and Technology, Tsinghua University, Beijing, China \\
  $^3$ Damo Academy, Alibaba Group, Hangzhou, China \\
  {\small\texttt{jiaofangkai@hotmail.com}}
  \quad {\small\texttt{guoyang.eric@gmail.com}}
  \quad {\small\texttt{niuyl14j@gmail.com}}\\
  {\small\texttt{\{zhongxiu.jf, fenglin.lfl\}@alibaba-inc.com}} \quad {\small\texttt{nieliqiang@gmail.com}}
}

\date{}

\begin{document}
\maketitle
\renewcommand{\thefootnote}{\fnsymbol{footnote}}
\footnotetext[1]{Work is done during internship at Alibaba Group.}
\footnotetext[2]{Corresponding author: Liqiang Nie.}
\renewcommand{\thefootnote}{\arabic{footnote}}
\begin{abstract}
Pre-trained Language Models (PLMs) have achieved great success on Machine Reading Comprehension (MRC) over the past few years. Although the general language representation learned from large-scale corpora does benefit MRC,
the poor support in evidence extraction which requires reasoning across multiple sentences hinders PLMs from further advancing MRC.
To bridge the gap between general PLMs and MRC, we present REPT, a {RE}trieval-based {P}re-{T}raining approach. In particular, we introduce two self-supervised tasks to strengthen evidence extraction during pre-training, 
which is further inherited by downstream MRC tasks through the consistent retrieval operation and model architecture.
To evaluate our proposed method, we conduct extensive experiments on five MRC datasets that require collecting evidence from and reasoning across multiple sentences.
Experimental results demonstrate the effectiveness of our pre-training approach.
Moreover, further analysis shows that our approach is able to enhance the capacity of evidence extraction without explicit supervision.\footnote{Our code and pre-trained models will be released at 
\href{https://github.com/SparkJiao/retrieval-based-mrc-pretraining}{github.com/SparkJiao/retrieval-based-mrc-pretraining}.}
\end{abstract}

\section{Introduction}

%
%

Machine Reading Comprehension (MRC) is an important task to evaluate the machine understanding of natural language.
Given a set of documents and a question (with possible options), an MRC system is required to provide the correct answer by either retrieving a meaningful span \cite{squad} or selecting the correct option from a few candidates \citep{race, dream, vqa1, adaVQA}.
Recently, with the development of self-supervised learning, the pre-trained language models \citep{bert,XLNet} fine-tuned on several machine reading comprehension benchmarks \citep{coqa,natural-questions} have achieved superior performance. 
The dominant reason lies in the strong and general contextual representation learned from large-scale natural language corpora.
Nevertheless, PLMs focus more on the general language representation and semantics to benefit various downstream tasks,
while MRC demands the capability of
extracting evidence across one or multiple documents and performing reasoning over the collected clues \cite{hgn,hotpotqa}.
Put it differently, there exists an obvious gap, indicating an insufficient exploitation of PLMs over MRC.

Some efforts have been made to bridge the gap between PLMs and downstream tasks, which can be roughly divided into two categories: knowledge enhancement and task-oriented pre-training \cite{ptm-surver-qiu}.
The former introduces commonsense or world knowledge into the pre-training \citep{ernie-thu,ernie2,mention-pretrain,coref-reasoning} or fine-tuning \cite{knowledge-mrc} for better performance over knowledge-driven tasks.
And the latter includes some delicately designed pre-training tasks,
e.g., the contrastive approach of learning discourse knowledge towards textual entailment task \cite{google-discourse-pretraining}.
Although these approaches have achieved some improvements on certain tasks, 
few of them are specifically designed for evidence extraction, which is indeed indispensable to MRC.

In fact, equipping PLMs with the capability of evidence extraction in MRC is challenging due to the following two factors.
1) The process of collecting clues from a document is difficult to be integrated into PLMs without designing specific model architectures or pre-training tasks \cite{ptm-surver-qiu,transformer-xh}. 
And 2) large-scale pre-training process would make PLMs overfit to pre-training tasks \cite{embedding-decoupling,transfer_language_model}. In other words, it is difficult to take full advantage of the pre-training merits if the training objectives of pre-training and downstream MRC are greatly separated.

\begin{figure}[t]
    \centering
    \includegraphics[width=0.48\textwidth]{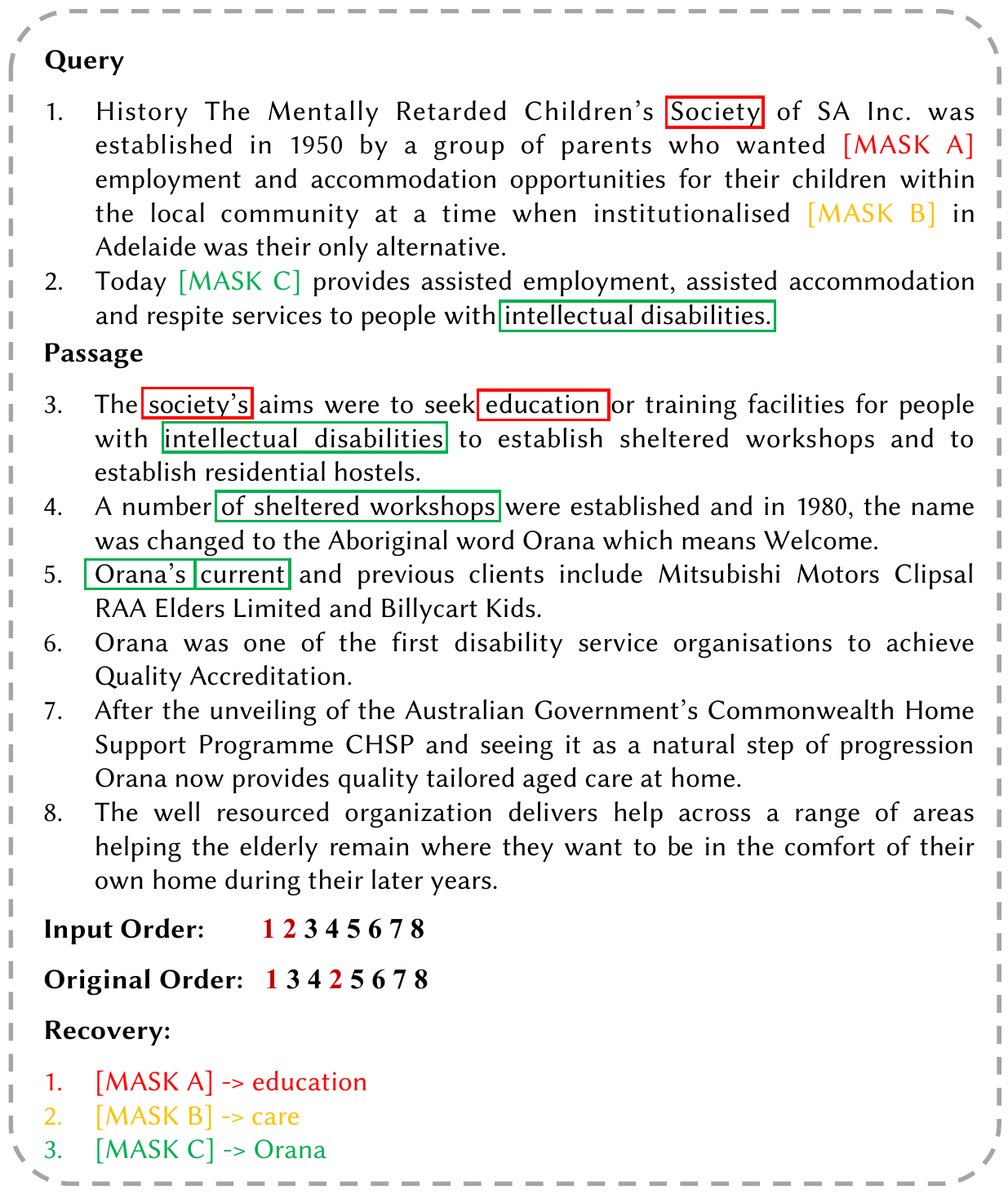}
    \caption{A running example obtained from our method. The query sentences are extracted from the original document with some crucial information being randomly masked, i.e., the sentence 1 and 2. The model is required to predict the preceding and following sentence for each query in the original document and recover the masked clues, i.e., infer the original order from input order and fill the [MASK] with the initial token. The phrases in boxes are the possible clues for recovering the masked tokens and the correct order.}
    \label{fig:example1}
    \vspace{-0.5cm}
\end{figure}

To deal with the aforementioned challenges, we propose a novel retrieval-based pre-training approach, REPT, to bridge the gap between PLMs and MRC. 
Firstly, to unify the training objective, we design a novel pre-training task, namely Surrounding Sentences Prediction (SSP), as illustrated in Figure~\ref{fig:example1}. Given a document, several sentences will be firstly selected as queries, and the others are jointly treated as a passage\footnote{We use \textit{passage} here to keep consistent with MRC tasks. And \textit{document} refers to the combination of queries and \textit{passage}.}.
Thereafter, for each query, the model should predict its preceding and following sentences in the original document by collecting clues from each sentence, which is compatible with evidence extraction in MRC tasks.
It is worth emphasizing that, the repeated occurrence of entities or nouns across different sentences often lead to information short-cut \cite{sentence-shuffling-emnlp}, from which the order of sentences can be easily recovered.
In view of this, we propose to mask such explicit clues.
As a result, the model is enforced to infer the correct positions of queries by gathering evidence with the incomplete information.
Secondly, to preserve the effectiveness of contextual representation, the masked clues are also required to be recovered through retrieving relevant information from other parts of the document, which is implemented via our Retrieval based Masked Language Modeling (RMLM) task.

In this way, the pre-training stage can be properly aligned with MRC:
1) the training objectives are connected through the introduction of the two pre-training tasks, which will be inherited by downstream MRC tasks through consistent retrieval operation.
And 2) the capability of evidence extraction from documents or sentences is enhanced during pre-training, and will be smoothly transferred to MRC.
Our contributions in this paper are summarized as follows:
\begin{enumerate}
    \item We present REPT, a novel pre-training approach, to bridge the gap between PLMs and MRC through retrieval-based pre-training.
    \item We design two self-supervised pre-training tasks, i.e., SSP and RMLM, to augment PLMs with the ability of evidence extraction with the help of retrieval operation and eliminating information short-cut, which can be smoothly transferred to downstream MRC tasks.
    \item We evaluate our method over five reading comprehension benchmarks of two different task forms: Multiple Choice QA (MCQA) and Span Extraction (SE). The substantial improvements over strong baselines demonstrate the effectiveness of our pre-training approach. We conduct an empirical study to verify that our method are able to enhance evidence extraction as expected.
\end{enumerate}


\section{Related Work}
MRC has received increasing attention in recent years. Many challenging benchmarks have been established to examine various forms of reasoning abilities, e.g., multi-hop \cite{hotpotqa}, discrete \cite{drop}, and logic reasoning \cite{reclor}. 
To solve the problem, a typical design is to gather possible clues through entity linking \cite{transformer-xh} or self-constructed graph \cite{hgn,num-net}, and then perform  multi-step reasoning. 
It is worth noting that, gathering clues is vital but challenging, especially for long document understanding. Some efforts have been dedicated to improving evidence extraction via direct \cite{r3-mrc} or distant supervision \cite{self-training-mrc}.

Generally, the fine-tuned PLMs \cite{bert,XLNet} can obtain superior performance in MRC due to their strong and general language representation. However, there still exist some gaps between PLMs and various downstream tasks, since certain abilities required by the downstream tasks cannot be learned through the existing pre-training tasks \cite{ptm-surver-qiu}. 
In order to take full advantage of PLMs, a few studies attempt to align the pre-training and fine-tuning stages.
For example, \citet{pretrainig_is_almost_all_you_need} reformulated the commonsense question answering task as scoring via leveraging the predicted probabilities for Masked Language Modeling (MLM) in RoBERTa \cite{roberta}. With the help of the commonsense learned through MLM, the method achieves comparable results with supervised approaches in zero-shot setting, indicating that bridging the gap between these two stages yields considerable improvement.
\citet{embedding-decoupling} tried to address the overfitting problem during pre-training through decoupling input and output embedding weights and enlarging the embedding size during decoding. The resultant model is therefore more transferable across tasks and languages.

In addition, some task-oriented pre-training methods have also been developed.
For instance, ~\citet{cross_thought_sentence_pretraining} proposed a novel pre-training method for sentence representation learning, where the masked tokens in a sentence are forced to be recovered from other sentences through sentence-level attention. Based on this, the attention weights can be directly fine-tuned to rank the candidates in answer selection or information retrieval.
\citet{odqa} tried to learn the dense document representation for information retrieval by minimizing the distance between the representation of an query sentence and its context.
\citet{realm} designed an augmented MLM tasks to jointly train a neural retriever and a language model for Open-domain QA. 
Different from these methods ranking the documents for open-domain QA, our approach focuses on enhancing the ability of evidence extraction in MRC, where the MLM based task by it alone is insufficient.


\section{Method}
\begin{figure*}[thbp]
    \centering
    \includegraphics[width=0.9\textwidth]{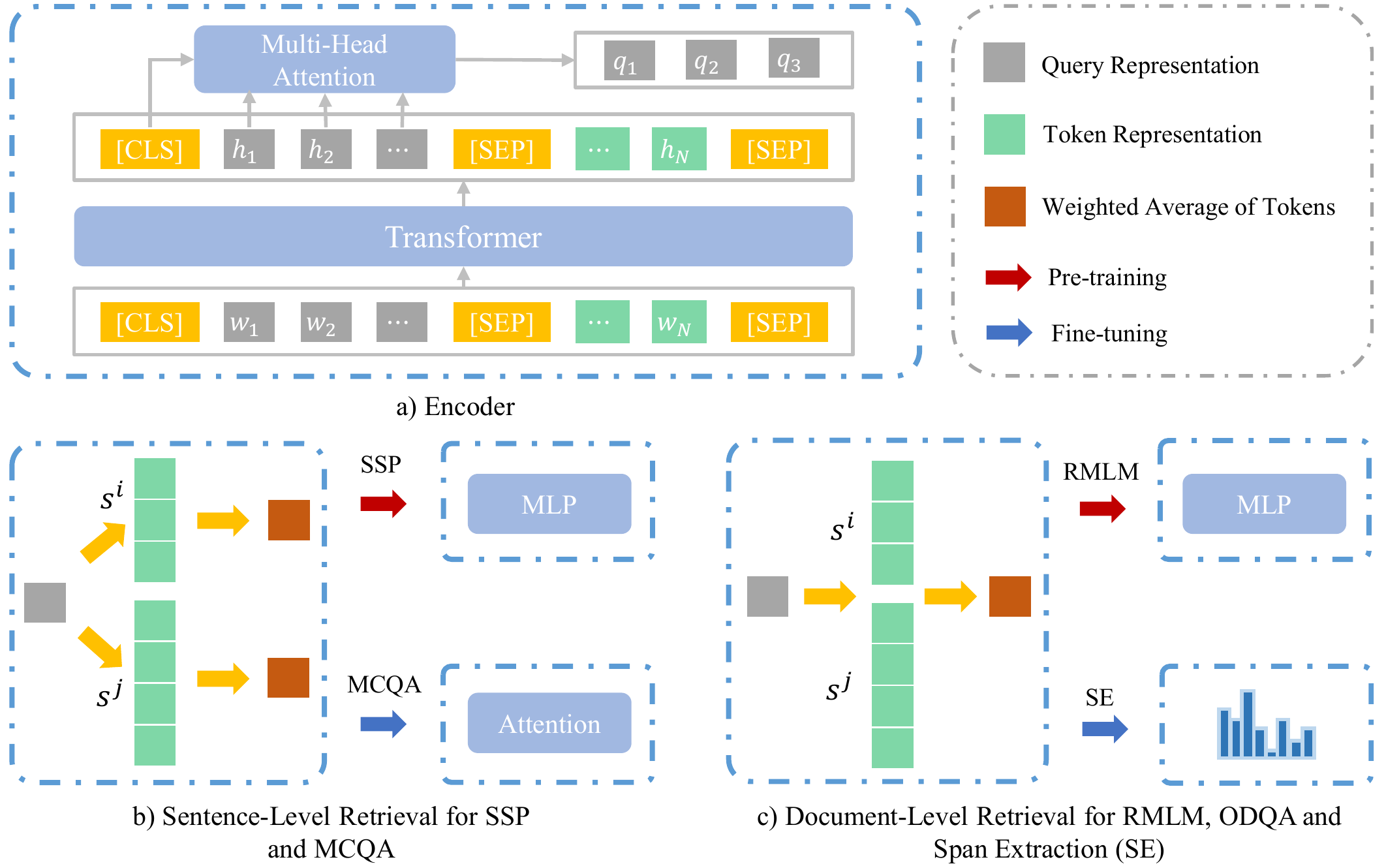}
    \caption{Framework of our model. a) Encoder composed of a pre-trained Transformer encoder and a query generator based on multi-head attention. b) The attention-based sentence-level retrieval for evidence extraction for each sentence, which will be further adopted by SSP during pre-training and MCQA during fine-tuning. c) The attention-based document-level retrieval for evidence extraction among the input sequence, which is employed for RMLM. For SE, the similarity function is directly fine-tuned. 
    }
    \label{fig:model-framework}
    \vspace{-0.4cm}
\end{figure*}

In this section, we present the details of the proposed method, \mtd.
We firstly describe the data pre-processing part (\S\ref{data-processing}), and then illustrate the two pre-training tasks, i.e., SSP and RMLM (\S\ref{model}) and the training objectives (\S\ref{optimization}).
Finally, we detail how to fine-tune our pre-trained model for downstream tasks through retrieval-based evidence extraction (\S\ref{fine-tuning}).


\subsection{Data Pre-processing}
\label{data-processing}

For pre-training, we use the English Wikipedia\footnote{We use the 2020/05/01 dump.} as our training data. We divide each Wikipedia article into segments, each containing up to 500 tokens\footnote{The tokenized sub-words following BERT and RoBERTa.} without overlapping. We treat each segment as a document and split it into several sentences\footnote{Any sentence with less than five tokens is concatenated to 
its previous one.}.

In order to increase the difficulty and efficiency of pre-training, for each document, we select 30\% of the most important sentences as queries and the rest in their original order as a passage. 
Specifically, the importance of each sentence in a document is measured through the summation of the importance of entities and nouns it contains, which is further defined as the number of sentences an entity/noun occurs.
Hereafter, masking is introduced to entities and nouns in queries according to pre-defined ratios to eliminate information short-cut. 
More details about the masking strategy are described in Appendix \ref{implementation-detail} and
an example after pre-processing can be found in Figure~\ref{fig:example1}.

\subsection{Task Definition}
\label{task-definition}
We treat a document as a sequence of $n$ sequential sentences with $m$ tokens. Supposing that there are $t$ sentences selected as queries following~\S\ref{data-processing}, the rearranged sequence is defined as $\mathcal{S}=[\mb{s}^1,\mb{s}^2,\cdots,\mb{s}^t, \cdots, \mb{s}^n]$, and the index of queries is $\mathcal{Q}=\{1,2,\cdots,t\}$. Besides, we define a mapping function $r$ to map the rearranged sentences to their original position. Taking Figure \ref{fig:example1} as an example, 
the mapping $r(\mb{s}^1)=1,\;r(\mb{s}^2)=4,\;r(\mb{s}^3)=2$ and $(\mb{s}^4)=3$ indicates that the original order is $\{\mb{s}^1,\mb{s}^3,\mb{s}^4,\mb{s}^2, \cdots\}$.

Taking $\mathcal{S}$ as input, the Surrounding Sentences Prediction task should predict the correct sentence index $a$ and $b$ for each query $\mb{s}^{q}$ with $q\in\mathcal{Q}$\footnote{Specifically, for $r(\mb{s}^q)=1$ or $r(\mb{s}^q)=n$, 
the corresponding prediction task is removed
since its preceding or following sentence does not exist.}:
\begin{equation}
    \left\{\begin{array}{l}
 r(\mb{s}^a)=r(\mb{s}^q)-1, \\ 
 r(\mb{s}^b)=r(\mb{s}^q)+1.
\end{array}\right.
\label{eq:restriction}
\end{equation}

As for the Retrieval based Masked Language Modeling (RMLM) task, the model should recover all the masked tokens in each query $\mb{s}^{q}$.

\subsection{Model}
\label{model}

First of all, we leverage a pre-trained Transformer \citep{transformer}, such as BERT, as our encoder to obtain the contextual representation of sentences.
The output of Transformer is formulated as:
\vspace{-0.4cm}
\begin{equation}
    \bh=[\mathbf{h}_{\bertcls}, \cdots,\mathbf{h}_{m},\mathbf{h}_{\bertsep}] = \mathrm{Encoder}(\mathcal{\tilde{S}}),
\end{equation}
where $\bh\in\mathbb{R}^{d\times (m+3)}$, and $d$ is the hidden size.
For a better illustration, we will use $\bh^{i}$ to represent the hidden state matrix of tokens that belong to sentence $\mb{s}^{i}$, 
such that:
\vspace{-0.2cm}
$$\bh=[\bh^{1},\bh^{2},\cdots,\bh^{n}],\;\bh^i\in\mathbb{R}^{d\times l_i},$$
where $l_i$ is the length of sentence $\mb{s}^i$ and $m=\sum_i l_i$. Since the process for each query is exactly the same, 
we use $q\in\mathcal{Q}$ as a representative to introduce the calculation with respect to each query below.

\subsubsection{Query Representation}
\label{query-representation-learning}
In order to gather potential clues from a document or sentences, we adopt the multi-head attention mechanism proposed by~\cite{transformer} to obtain the sentence-level representation for each query.
Formally, the attention mechanism is defined as $\mathrm{MHA}(\mathbf{Q},\mathbf{K}, \mathbf{V})$, where $\mathbf{Q},\mathbf{K},\mathbf{V}$ are query, key and value matrices, respectively. To consider the global information, we leverage $\mathbf{h}_{\bertcls}$ as the query vector, and $\bh^{q}$ as $\mb{K}$ and $\mb{V}$: 
\begin{equation}
    {\mathbf{v}^{q}_0}^\top=\mathrm{MHA}(\mathbf{h}_{\bertcls}^\top, \bh^{q}, \bh^q).
    \label{query1}
\end{equation}

During pre-training, we reuse the layer defined by Equation \ref{query1} with $\mb{Q}=\mb{v}^q_0$ and $\mb{K}=\mb{V}=\bh^q$, to generate the task-specific query representation $\mb{v}^q$, which is designed to alleviate the overfitting problem \citep{he2020deberta}.

\subsubsection{Surrounding Sentence Prediction}

To enhance the capability of pre-trained models for evidence extraction, we have carefully designed the SSP task, where the model should predict the preceding and following sentences for a given query by extracting the relevant evidence from each sentence.
Consequently, we introduce a retrieval operation, which is implemented via a single-head attention mechanism\footnote{The details are illustrated in Appendix~\ref{sec:simple-attention}.}:

\begin{equation}
    {\mathbf{u}^{i}_q}^\top=\mathrm{Att}({\mathbf{v}^{q}}^\top, \bh^{i}, \bh^{i}),
    \label{sen-sum}
\end{equation}
where $\mathbf{u}^{i}_q$ is the representation of sentence $\mb{s}^{i}$, highlighting the evidence information pertaining to query $\mb{s}^{q}$. Finally, the score of each sentence in the document with regard to $\mb{s}^{q}$ is obtained through:
\begin{equation}
    {\mathbf{o}^{i}_q}=\mathbf{W}_2(\tanh(\mathbf{W}_1 {\mathbf{u}^{i}_q}+\mathbf{b}_1))+\mathbf{b}_2.
\end{equation}
\vspace{-0.1cm}

\subsubsection{Retrieval based MLM}

Since the masking noise introduced when constructing queries could also bring inconsistency between pre-training and fine-tuning, we further designed a retrieval based MLM task to alleviate this problem. In the RMLM task, the model should predict the masked entities or nouns through retrieving relevant information from a document. 
More specifically, the query-aware evidence representation of the input sequence is obtained via:
\begin{equation}
    {\mathbf{g}^{q}}^\top=\mathrm{Att}({\mathbf{v}^{q}}^\top, \bh, \bh).\label{doc-sum}
\end{equation}
Denoting the index of a masked token in query $\mb{s}^q$ as $z$, the representation of the masked token $s^{q}_{z}$ used for recovering is:
\begin{equation}
    \tilde{\mathbf{h}}_{z}^{q}=f(\mathbf{h}_{z},\mathbf{g}^{q}),
    \label{eq:rmlm-predict}
\end{equation}
where the function $f(\cdot,\cdot)$ is implemented as a normalized 2-layer feed-forward network, and the details are illustrated in Appendix \ref{ff-gleu}.




\subsection{Optimization}
\label{optimization}

As the definition in Equation~\ref{eq:restriction}, given $a$ and $b$ as the index of the original preceding and following sentences of the query $\mb{s}^q$ in $\mathcal{S}$, 
the corresponding probabilities for surrounding sentences are formulated as:
\begin{equation}
\left\{
    \begin{aligned}
        p_{\mathrm{ssp}}(a|q,\mathcal{S})&=\frac{\exp(\mathbf{o}^a_q)}{\sum^n_{j=1,j\notin\{b,q\}}\exp(\mathbf{o}^j_q)}, \\
        p_{\mathrm{ssp}}(b|q,\mathcal{S})&=\frac{\exp(\mathbf{o}^b_q)}{\sum^n_{j=1,j\notin\{a,q\}}\exp(\mathbf{o}^j_q)}.
    \end{aligned}
    \right.
    \label{ssp-prob}
\end{equation}
The objective of SSP is subsequently defined as:
\begin{equation}
\begin{aligned}
    \mathcal{L}_{\mathrm{ssp}}=\mathbb{E}(-\frac{1}{|\mathcal{Q}|}\sum_{q}(&\log{p_{\mathrm{ssp}}(a|q,\mathcal{S})}+ \\
    &\log{p_{\mathrm{ssp}}(b|q,\mathcal{S})})).
\end{aligned}
\end{equation}
As for RMLM, supposing the index set of masked tokens in query $\mb{s}^q$ is $\mathcal{Z}^q$, and the set of corresponding original tokens is $\mathcal{X}^q$, the probability for recovering a masked token is:
\begin{equation}
    p_{\mathrm{rmlm}}(x_z|z,q,\mathcal{S})=\frac{\exp(\mathrm{e}(x_z)^\top \tilde{\mb{h}}_{z}^{q})}{\sum_{x'} \exp(\mathrm{e}(x')^\top \tilde{\mb{h}}_{z}^{q})},
\end{equation}
where $z\in\mathcal{Z}^q,x_z\in\mathcal{X}^q$, $x'$ is a token in vocabulary, and $\mathrm{e}(x)$ denotes the word embedding of $x$. 
Then the objective of RMLM is:
\begin{equation}
    \begin{aligned}
        \mathcal{L}_{\mathrm{rmlm}}=\mathbb{E}(
            -\frac{\sum_{q}\sum_{z}\log p_{\mathrm{pmlm}}(x_z|z,q,\mathcal{S})}{\sum_{q}|\mathcal{Z}^q|}
        ).
    \end{aligned}
\end{equation}

During pre-training, the model tries to optimize the two objectives jointly:
\begin{equation}
    \mathcal{L}=\mathcal{L}_\mathrm{ssp}+\mathcal{L}_\mathrm{rmlm}.
\end{equation}

\subsection{Fine-tuning}
\label{fine-tuning}

During fine-tuning, the input contains a query sentence and a passage. For multiple choice QA tasks, we concatenate a question with an option to form a question-option pair and use it as a whole query. In this section, we use $q=0$ to represent the index of the query and the sentences of passage are kept in their original order. The input sequence can be thus denoted as:
$$
\mathcal{S}=[s^q,s^1,s^2,\cdots,s^n].
$$

To inherit the evidence extraction ability augmented during pre-training, we incorporate the same retrieval operation into fine-tuning to collect clues from the passage. Firstly, we reuse the attention mechanism defined in Equation \ref{query1} to obtain the query representation $\mb{v}^q$.
As for the evidence extraction process, we formulate it differently for Multiple Choice QA and Span Extraction.

\subsubsection{Multiple Choice QA}
\label{fine-tuning-mcqa}

Similar to Equation \ref{sen-sum}, we adopt an attention mechanism, whereby the query-aware sentence representation $\mb{u}^i_q$ is obtained via gathering evidence from each sentence:
\begin{equation}
    {\mathbf{u}^i_q}^\top=\mathrm{Att}({\mb{v}^q}^\top, \bh^i, \bh^i),\;i\neq q.
\end{equation}
And the final passage representation highlighting the evidence can be obtained via the sentence-level evidence extraction:
\begin{equation}
    \mathbf{v}^p=\mathrm{Att}({\mathbf{v}^q}^\top, \mb{U}, \mb{U}),
\end{equation}
where $\mb{U}=[\mb{u}^1_q,\cdots,\mb{u}^n_q]$ and $\mb{U}\in\mathbb{R}^{d\times n}$.
Finally, we represent the probability of each option $c$ using both the query $\mathbf{v}^q$ and the passage $\mathbf{v}^p$:
\begin{equation}
    p^{\mathrm{mc}}_c\propto \exp(\mathbf{W}_6(\tanh(\mathbf{W}_5[\mathbf{v}^q;\mathbf{v}^p]+\mathbf{b}_5))+\mathbf{b}_6).
\end{equation}

Specifically, for Multi-RC, since the number of correct answer options for each question is uncertain, the task is often treated as a binary classification problem for each option. As a result, we adopt a MLP to get the probability of whether an option $c$ is correct:
\begin{equation}
    p^{\mathrm{mc}}_c = \sigma(\mathbf{W}_8(\tanh(\mathbf{W}_7[\mathbf{v}^q;\mathbf{v}^p]+\mathbf{b}_7))+\mathbf{b}_8),
\end{equation}
where $\sigma$ is the $\mathrm{sigmoid}$ function.

\subsubsection{Span Extraction}
\label{span-extraction}
Since answer spans are often consistent with corresponding evidences, we directly leverage the query to extract relevant spans. The probability of selecting start position $s$ and end position $e$ of an answer span is given by:
\begin{equation}
\left\{
    \begin{aligned}
        p^{\mathrm{span}}_s\propto\exp({\mathbf{v}^q}^\top \mathbf{W}_9 \mathbf{h}_{s}), \\
        p^{\mathrm{span}}_e\propto\exp({\mathbf{v}^q}^\top \mathbf{W}_{10} \mathbf{h}_{e}).
    \end{aligned}
\right.
\end{equation}



\section{Experiment}

\subsection{Dataset}

\begin{table*}[t]
\centering
\setlength{\tabcolsep}{3.2mm}{
\scalebox{0.9}{
\begin{tabular}{l|cc|cc|cc|ccc}
\toprule
             & \multicolumn{2}{c}{\textbf{RACE}} & \multicolumn{2}{|c}{\textbf{DREAM}} & \multicolumn{2}{|c}{\textbf{ReClor}} & \multicolumn{3}{|c}{\textbf{Multi-RC}} \\
Model / Dataset & Dev       & Test        & Dev         & Test        & Dev         & Test          & \multicolumn{3}{c}{Dev} \\
             & Acc.      & Acc.           & Acc.        & Acc.          & Acc.        & Acc.          & EM      & F1$_a$      & F1$_m$ \\ \hline
BERT-base\dag &  --       & 65.0           & 63.4        & 63.2          &\textbf{54.6}& 47.3          & --      & --        & --  \\
BERT w. M    & 67.7      & 66.3           & 62.9        & 63.2          & 51.6        & 45.1          & 26.6    & 71.8      & 74.2    \\
\modb        & 67.2      & 65.2           & 62.9        & 62.3          & 48.4        & 45.0          & 22.8    & 69.6      & 72.0   \\
\modb w. M   & 67.7      & 66.9           & 61.8        & 62.2          & 48.8        & 48.3          & 23.8    & 70.1      & 72.6    \\
\modb w. R   & 65.5      & 64.7           & 59.0        & 58.6          & 46.8        & 45.1          & 26.4    & 71.5      & 74.0    \\
\modb w. S   & 69.5      & 66.5           &\textbf{64.8}& 62.2          & 52.0        & 46.5          & 30.0    & 73.0      & 75.8   \\
\modb w. R/S &\textbf{70.1}&\textbf{68.1} &64.4         &\textbf{64.0}  & 50.6        &\textbf{49.2}  &\textbf{31.9}&\textbf{73.8}&\textbf{76.3}\\ \hline
RoBERTa-base & 76.0      & 75.5           & \textbf{71.2}& 69.8         & 54.8        & \textbf{50.8} & 38.7    & 77.1      & 79.2      \\
\modrb       & 76.8      & \textbf{75.7}  & 70.9        & 69.5          & \textbf{56.0}& 49.7         & 34.6    & 75.4      & 77.4      \\
\modrb w. R/S& \textbf{77.1} & 74.9       & 70.9        & \textbf{70.8} & 54.8        & 50.3          & \textbf{40.4}& \textbf{77.6} & \textbf{80.0} \\ \bottomrule
\end{tabular}
}}
\caption{Results on multiple choice question answering tasks. (F1$_a$: F1 score on all answer-options; F1$_m$: macro-average F1 score of all questions.) We ran all experiments using \textbf{four} different random seeds with the same hyper-parameters, and report the average performance, except for ReClor and Multi-RC. For ReClor, we submitted the best model on the development set to the leaderboard to get the results on the test set. For MultiRC, we merely reported the performance on development set since the test set is unavailable. $\dag$: The results are reported by the leaderboard.}
\vspace{-0.3cm}
\label{mcqa-results}
\end{table*}

\subsubsection{Multiple Choice Question Answering}

\textbf{DREAM}~\citep{dream} contains 10,197 multiple choice questions for 6,444 dialogues collected from English Examinations designed by human experts, in which 85\% of the questions require reasoning across multiple sentences, and 34\% of the questions also involve commonsense knowledge.

\noindent\textbf{RACE}~\citep{race} is a large-scale reading comprehension dataset collected from English Examinations and created by domain experts to test students' reading comprehension skills.
It has a wide variety of question types, e.g., summarization, inference, deduction and context matching, and requires complex reasoning techniques.


\noindent\textbf{Multi-RC}~\citep{multirc} is a dataset of short paragraphs and multi-sentence questions. The number of correct answer options for each question is not pre-specified and the correct answer(s) is not required to be a span in the text. 
Moreover, the dataset provides annotated evidence sentence. 


\noindent\textbf{ReClor}~\citep{reclor} is extracted from logical reasoning questions of standardized graduate admission examinations. 
Existing studies show that the state-of-the-art models perform poorly on ReClor, indicating the deficiency of logical reasoning ability of current PLMs.

\subsubsection{Span Extraction}

\textbf{Hotpot QA}~\citep{hotpotqa} is a question answering dataset involving natural and multi-hop questions. The challenge contains two settings, the distractor setting and the full-wiki setting. 
In this paper, we focused on the full-wiki setting, where the system should retrieve the relevant paragraphs from Wikipedia and then predict the answer.

\noindent\textbf{SQuAD2.0}~\cite{squad2.0} is reading comprehension dataset, consisting of questions posed by crowdworkers on a set of Wikipedia articles, where the answer to every question is a segment of text, or span, from the corresponding reading passage, or the question might be unanswerable.

\subsection{Implementation Detail}
We leave the details about the implementation and pre-training corpora in Appendix \ref{implementation-detail} due to the limitation of space.

\subsection{Baseline}

Since our method is used for further pre-training, we mainly compared our model with BERT/RoBERTa and their variants. For Hotpot QA, we integrated our models into an open-sourced and well-accepted system \cite{learning-to-retrieve} and evaluated the performance.
The details of baselines are summarized as follows:

\subsubsection{Multiple Choice QA}
\textbf{BERT} is the BERT-base model with 2-layer MLP as the task-specific module.

\noindent\textbf{\modb}\& \textbf{\modrb} refer to the designed but not further trained models, which include an extra multi-head attention for generating query representation via Equation \ref{query1}, and our retrieval operation for evidence extraction as in \S\ref{fine-tuning-mcqa} and \S\ref{span-extraction}. 

\noindent\textbf{\modb w. R/S} \& \textbf{\modrb w. R/S} refer to the designed models further trained with our proposed SSP and RMLM tasks (denoted as \textbf{S} and \textbf{R}, respectively). 

\noindent\textbf{\modb w. R} \& \textbf{\modb w. S} refer to the models further trained with only one pre-training task, RMLM or SSP.


\noindent\textbf{\modb w. M} \& \textbf{BERT w. M} refer to the models further trained with MLM. For fair comparison, we further train BERT with the same Wikipedia corpus for equivalent steps. 

\subsubsection{Hotpot QA}

For hotpot QA, we constructed the system based on Graph-based Recurrent Retriever \cite{learning-to-retrieve}, which includes a retriever and a reader based on BERT. We simply replaced the reader with our models and evaluated their performance in comparison with several published strong baselines on the leaderboard\footnote{https://hotpotqa.github.io/.}.

\section{Results and Analyses}


\subsection{Results for Multiple Choice QA}

Table \ref{mcqa-results} shows the results of the baselines and our method on multiple choice question answering. 


From Table \ref{mcqa-results}, we can observe that: 
1) Compared with \modb and BERT, our method significantly improves the performance over all the datasets, which validates the effectiveness of our proposed pre-training method.
2) As for the model structure, \modb obtains similar or worse results compared with BERT, which suggests that the retrieval operation can hardly improve the performance without specialised pre-training.
3) Taking the rows of BERT, BERT-Q, BERT w. M, \modb w. M for comparison, the models with further pre-training using MLM achieve similar or slightly higher performance.
The results show that further training BERT using MLM and the same corpus can only achieve very limited improvements.
4) Regarding the two pre-training tasks, \modb w. R/S leads to similar performance on the development sets compared with \modb w. S, but a much higher accuracy on the test sets, which suggests RMLM can help to maintain the effectiveness of contextual language representation.
However, there is a significant degradation over all datasets for \modb w. R. The main reason is possibly because the model cannot tolerate the sentence shuffling noise, which may lead to the discrepancy between pre-training and MRC, and thus need to be alleviated through SSP.
And 5) considering the experiments over RoBERTa-based models, \modrb w. R/S outperforms \modrb and RoBERTa-base with considerable improvements over Multi-RC and the test set of DREAM, which also indicates that our method can benefit stronger PLMs.


\subsection{Performance on Span Extraction QA}

The results of span extraction on Hotpot QA are shown in Table \ref{tab:hotpot-results}. 
We constructed the system using the Graph Recurrent Retriever (GRR) proposed by \citet{learning-to-retrieve} and different readers.
As shown in the table, GRR + \modb w. R/S outpeforms GRR + BERT-base by more than 2.5\% absolute points on both EM and F1. And GRR + \modrb w. R/S also achieves a significant improvement over GRR + RoBERTa-base. During the test stage, our best system, GRR + \modrb w. R/S performs better than the strong baselines and get closer to GRR + BERT-wwm-large.
The above results strongly demonstrate the effectiveness of our pre-training method on the task requiring multi-hop evidence extraction and reasoning.

Besides, we also conducted experiments on the most common benchmark, SQuAD2.0. The results on development set shown in Table~\ref{tab:squad-results} have also verified the effectiveness of our proposed pre-training method.

\begin{table}[t]
\centering
\scalebox{0.75}{
\begin{tabular}{l|cc|cc}
\toprule
\multicolumn{1}{c|}{Model / Dataset}  & \multicolumn{2}{c|}{Dev}                & \multicolumn{2}{c}{Test} \\
                     & EM               & F1     & EM       & F1     \\ \hline
Transformer-XH \cite{transformer-xh}       & 54.0                & 66.2      & 51.6  & 64.7  \\
HGN \cite{hgn}                  & --                  & --        & 56.7  & 69.2 \\ 
GRR + BERT-wwm-Large*        & \textbf{60.5}                  & \textbf{73.3}      & \textbf{60.0}        & \textbf{73.0}        \\ \hline
GRR + BERT-base*       & 52.7                & 65.8      & --          & --         \\
GRR + \modb w. R/S    & \textbf{55.2}       & \textbf{68.4}     & --          & --         \\ \hline
GRR + RoBERTa-base    & 56.8                  & 69.6      & --          & --          \\
GRR + \modrb w. R/S   & \textbf{58.4}            & \textbf{71.3}     & 58.1          & 71.0          \\ \bottomrule
\end{tabular}
}
\caption{Results of our method and other strong baselines on Hotpot QA. \textit{GRR} means the Graph Recurrent Retriever proposed by \citet{learning-to-retrieve}, \textit{GRR + BERT-base} means the system whose retriever is GRR and reader is built on BERT-base. *:  The results are reported by \citet{learning-to-retrieve}.}
\vspace{-0.2cm}
\label{tab:hotpot-results}
\end{table}

\begin{table}[t]
\centering
\scalebox{1.0}{
\begin{tabular}{l|cc}
\toprule
Model / Dataset  & EM             & F1    \\ \hline 
\modb            & 71.7           & 74.9          \\
\modb w. R/S     & \textbf{77.2}  & \textbf{80.4}             \\ \hline
\modrb           & 80.3           & 83.7              \\
\modrb w. R/S    & \textbf{81.7}  & \textbf{85.0}         \\ \bottomrule
\end{tabular}
}
\caption{Results of our method and other baselines on the dev set of  SQuAD2.0.}
\vspace{-0.2cm}
\label{tab:squad-results}
\end{table}




\subsection{Evaluation of Evidence Extraction}
\label{sub-sec:evidence-extraction}

To evaluate the performance of our method for evidence extraction in the setting of implicit supervision (with only answers), we ranked sentences in a passage using their attention weights obtained in Equation~\ref{sen-sum} and chose those sentences with higher weights as the evidences.

As shown in Table \ref{tab:multirc-evidence}, the models with our proposed pre-training tasks obtain considerable improvements on the precision and recall of evidence extraction, which verifies that our pre-training method is able to effectively equip PLMs with the capability for gathering evidence without explicit supervision. For a better illustration, we further provided two examples in Appendix \ref{sec:case-study}. 





\begin{table}[t]
\centering
\scalebox{0.85}{
\begin{tabular}{l|cc|cc}
\toprule
Model          & \textbf{P@1} & \textbf{R@1} & \textbf{P@2} & \textbf{R@2} \\ \hline
\modb           & 21.83              & 9.66             & 20.24          & 17.73 \\
\modb w. R/S    & \textbf{45.30}     & \textbf{20.38}   & \textbf{38.51} & \textbf{34.55} \\ \hline
\modrb        & 28.25     & 12.45   & 26.93 & 23.74 \\
\modrb w. R/S & \textbf{35.34}     & \textbf{15.76}   & \textbf{30.33} & \textbf{26.85}  \\ \bottomrule
\end{tabular}
}
\caption{Results of evidence extraction on the development set of Multi-RC.}
\label{tab:multirc-evidence}
\vspace{-0.2cm}
\end{table}

\begin{table}[t]
\scalebox{0.85}{
\centering
\begin{tabular}{l|cc|ccc}
\toprule
                 & \multicolumn{2}{c|}{RACE} & \multicolumn{3}{c}{Multi-RC} \\
Model/Dataset    & Dev         & Test        & \multicolumn{3}{c}{Dev}      \\
                 & Acc.        & Acc.        & EM       & F1$_a$      & F1$_m$     \\ \hline
B.Q w.R/S (30\%) & 70.1          & 68.1          & 31.9           & \textbf{73.8}   & \textbf{76.3}   \\
B.Q w.R/S (60\%) & 70.2          & 67.3          & \textbf{32.0}  & \textbf{73.8}   & \textbf{76.3}     \\
B.Q w.R/S (90\%) & \textbf{70.4} & \textbf{68.2} & 31.0           & 73.5            & 76.2    \\ \hline
B.Q w.S (No Mask)    & 69.0          & 67.2          & 29.0           & 72.7            & 75.4     \\\bottomrule
\end{tabular}
}
\caption{Results on RACE and Multi-RC using models pre-trained with different mask ratios. \textit{B.Q} means \textit{BERT-Q}.}
\label{tab:mask-ratio-ablation}
\vspace{-0.4cm}
\end{table}

\subsection{Effect of Different Masking Ratio During Pre-training}

Table~\ref{tab:mask-ratio-ablation} shows the results of our model pre-trained with different masking ratios. Due to the small amount of entities contained in the document, we only consisdered the masking ratio of nouns as the variable. Formally, we considered three ratios: 30\%, 60\%, 90\%, and an extra setting, where the entities and nouns are all kept and the RMLM task is also removed during pre-training.

As shown in the table, with more possible clues being masked, the model tend to obtain better results on the downstream tasks. For example, \modb w. R/S (90\%) achieves the best accuracy on RACE, and \modb w. R/S (60\%) obtains the highest performance over Multi-RC.
And all models that employ masking outperform \modb w. S (no masking). The main reason can be that with more explicit information short-cut being eliminated, it is more difficult for models to collect potential clues, and PLMs are enhanced with stronger reasoning ability of evidence extraction. 
However, there also exists a trade-off: as higher masking ratio leads to more noise, it could worsen the mismatch between pre-training and fine-tuning, and cause performance degradation, e.g., \modb w. R/S (90\%) performs the worst on Multi-RC. 


\begin{figure}[t]
    \centering
    \includegraphics[width=0.45\textwidth]{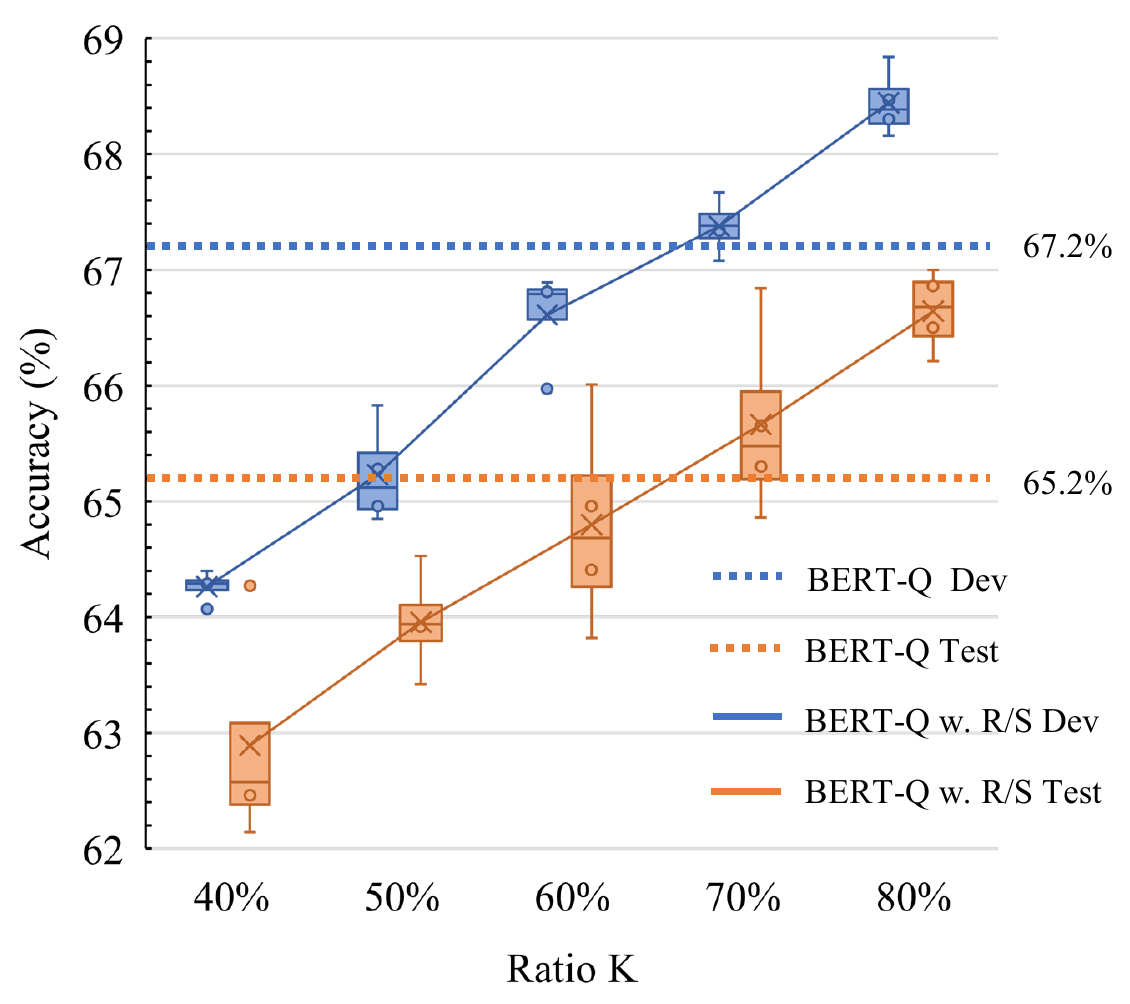}
    \caption{The accuracy of \modb w. R/S on the development and test of RACE. The horizontal axis refers to the ratio $K$ of training data compared to the original training set.}
    \label{fig:race_low_resouce}
    \vspace{-0.4cm}
\end{figure}

\subsection{Performance in Low Resource Scenario}

Figure~\ref{fig:race_low_resouce} depicts the performance of \modb w. R/S on the development and test set of RACE with limited training set. For each specific relative ratio, four reduced training sets are automatically generated using different random seeds and the corresponding accuracies are plotted on the figure.
It is observed that with 70\% training data, our model outperforms the baseline, BERT-Q, which was initialized using BERT and has not been further pre-trained. The results indicate that our method can help to reduce the amount of annotated training data for downstream MRC tasks, which is especially useful in low resource scenarios.



\section{Conclusion and Future Work}
In this paper, we present a novel pre-training approach, REPT, to bridge the gap between pre-trained language models and machine reading comprehension through retrieval-based pre-training. 
Specifically, we design two retrieval-based pre-training tasks equipped with self-supervised learning, namely Surrounding Sentences Prediction (SSP) and Retreval based Masked Language Modeling (RMLM), to enhance PLMs with the capability of evidence extraction for MRC. 
The experiments over five different datasets validate the effectiveness of our proposed method. In the future, we plan to extend the proposed pre-training approach to the more challenging open-domain settings.

\section{Acknowledgements}

This work is supported by the National Key Research and Development Project of New Generation Artificial Intelligence, No.:2018AAA0102502, and the Alibaba Research Intern Program of Alibaba Group.

\bibliographystyle{acl_natbib}

\clearpage

\appendix
\section{Implementation Detail}
\label{implementation-detail}

We built our model on Huggingface's Pytorch transformer repository~\citep{huggingface}, and used AdamW \citep{adamw} as the optimizer. We used the pre-trained BERT-base-uncased and RoBERTa-base checkpoint to initialize our encoder, and performed pre-training using 16 P100 GPUs simultaneously. The pre-training processes last around 16 hours for BERT and 4 days for RoBERTa, which takes 20,000 steps and 80,000 steps with the batch size as 512, respectively. All hyper-parameters can be found in Table \ref{tab:hyperparam-pretraining} for pre-training and Table \ref{tab:hyperparam-fine-tuning} for fine-tuning.

During constructing the training sample for pre-training, we controlled the masking ratio for entity and noun in query. For BERT, we masked 90\% entities and 30\% nouns. For RoBERTa, we constructed two datasets, where the masking ratios for entity and noun are set to 90\%, 30\% and 90\%, 90\%, respectively. And we mixed the two for jointly training. We also explored the effect of different masking ratios and the analysis is detailed in \S\ref{tab:mask-ratio-ablation}.

As for the fine-tuning stage, for multiple choice QA, we ran all experiments using for different random seeds (i.e., 33, 42, 57 and 67) and reported the average performance, except for ReClor, in which we only submitted the results obtained from the model which performs the best on development set to the leaderboard because the limitation of submission times. For Hotpot QA, we mainly followed the hyper-parameters of \citet{learning-to-retrieve} and thus did not repeat the experiments using different random seeds. Due to the submission limitation, we only submitted our best model on the development set to the leaderboard and reported its performance on test set.

\section{The Details About Modeling}
\subsection{Single-head Attention}
\label{sec:simple-attention}

To reduce the extra parameters introduced, we define a single-head attention mechanism compared to the multi-head one. Given the query matrix $\mb{Q}$, key matrix $\mb{K}$ and value matrix $\mb{V}$, the simple attention mechanism is formualted as:
$$
\begin{aligned}
\mathrm{Att}(\mb{Q}, \mb{K}, \mb{V})&= \mathrm{softmax}((\mb{Q}\mb{W}+\mb{b})^\top\mb{K})\mb{V},
\end{aligned}
$$
where $\mb{W}$ and $\mb{b}$ is the learnable parameters.

\subsection{Normalized Feed-forward Network}
\label{ff-gleu}
We adopt a 2-layer feed-forward network with GeLU activation \cite{gelu} and layer normalization \cite{layer_norm} to predict the masked entities and nouns. Following SpanBERT \citep{SpanBERT}, the Equation \ref{eq:rmlm-predict} is decomposed as:
$$
\begin{aligned}
    \left\{\begin{array}{l}
          \mathbf{h}_0=[\mathbf{h}_{z};\mathbf{g}^{q}],\\
          \mathbf{h}_1=\mathrm{LayerNorm}(\mathrm{GeLU}(\mathbf{W}_3\mathbf{h}_0+\mathbf{b}_3)),\\
          \tilde{\mathbf{h}}_{z}^{q}=\mathrm{LayerNorm}(\mathrm{GeLU}(\mathbf{W}_4\mathbf{h}_1+\mathbf{b}_4)).
    \end{array}\right.
\end{aligned}
$$

\section{Case Study About Evidence Extraction}
\label{sec:case-study}

\begin{figure*}[t]
    \centering
    \includegraphics[width=0.95\textwidth]{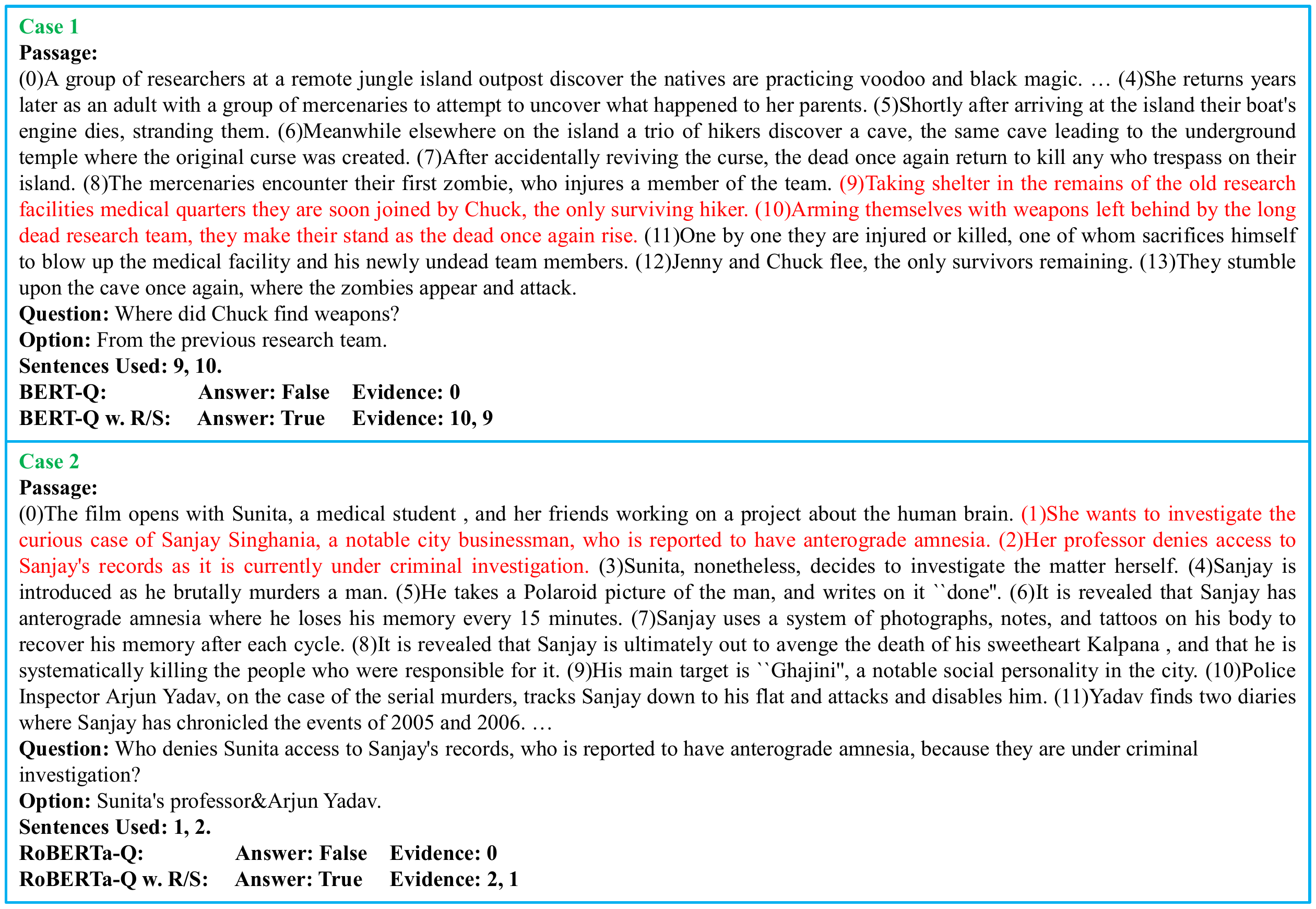}
    \caption{Two cases from the development set of Multi-RC.}
    \label{fig:case_study}
\end{figure*}

\begin{figure}[t]
    \centering
    \subfigure[Normalized attention weights for Case 1 in Figure 4.]{
        \label{fig:attn-weights-case1}
        \includegraphics[width=0.95\linewidth]{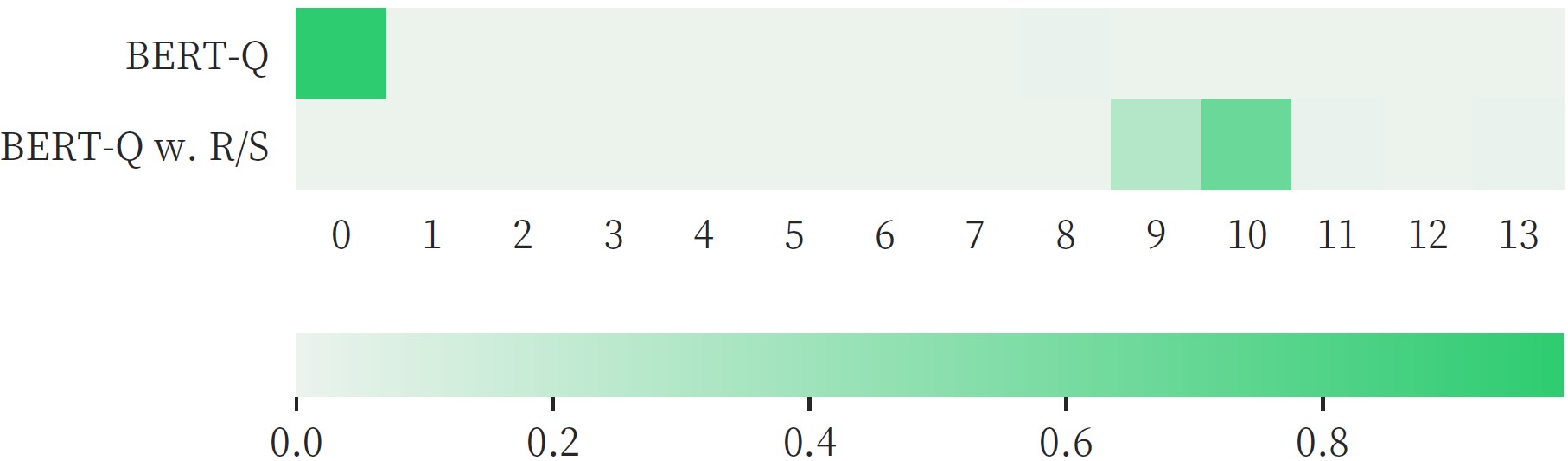}
    }
    
    \subfigure[Normalized attention weights for Case 2 in Figure 4.]{
        \label{fig:attn-weights-case2}
        \includegraphics[width=0.95\linewidth]{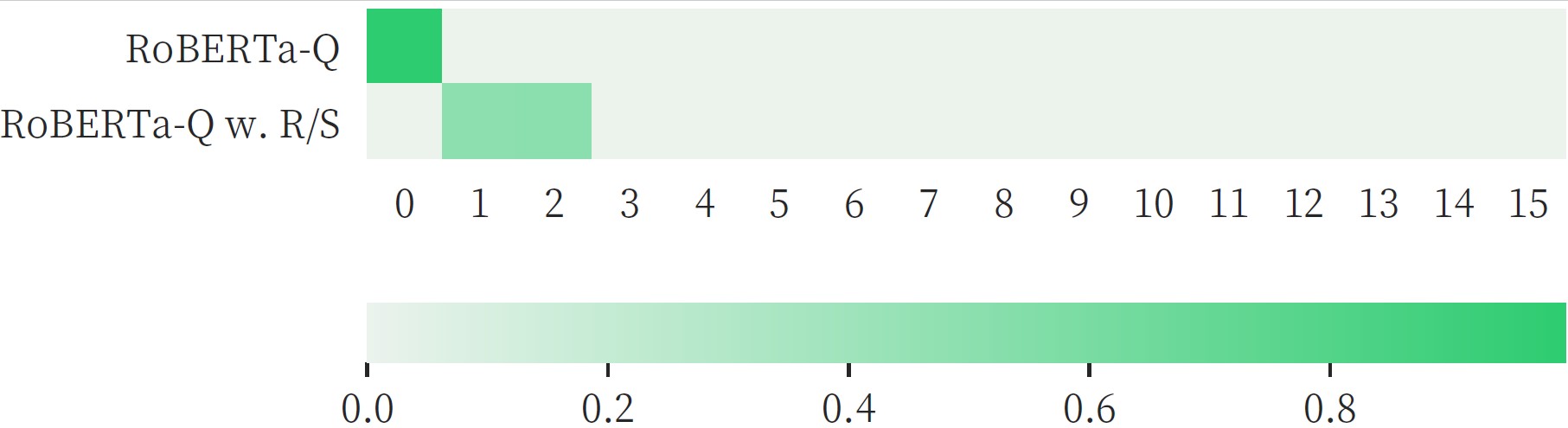}
    }
    
    \caption{Two cases of the normalized attention weights of evidence extraction.}
    \label{fig:attn-weights}
\end{figure}

In \S\ref{sub-sec:evidence-extraction}, the results show that our pre-training method can augment the ability to extract the correct evidence. To give an intuitive clarification over this, we select two cases shown in Figure~\ref{fig:case_study}. As we can see, \modb w. R/S and \modrb w. R/S can select the correct evidence sentences, while the baselines models attend to the wrong sentences. Besides, Figure~\ref{fig:attn-weights} shows the attention maps of the two groups of comparison. It can be observed that our pre-training approach can help the model learn a uniform attention distribution over the possible evidence sentences.



\begin{table*}[]
\centering
\scalebox{0.9}{
\begin{tabular}{lcccccc}
\toprule
\textbf{HyperParam}     & \textbf{BERT-base} & \textbf{RoBERTa-base} \\ \hline
Peak Learning Rate      &  2e-4              & 5e-5                             \\
Learning Rate Decay     &  Linear            & Linear                       \\
Batch Size              &  512               & 512                            \\
Max Steps               &  20,000            & 80,000                                  \\
Warmup Steps            &  2,000              & 4,000                                \\
Weight Decay            &  0.01              & 0.01                                   \\
Gradient Clipping       &  1.0               & 0.0                                   \\
Adam $\epsilon$         &  1e-6              & 1e-6                                   \\
Adam $\beta_1$          &  0.9               & 0.9                                  \\
Adam $\beta_2$          &  0.999             & 0.98                                   \\
Max Sequence Length     &  512               & 512                                  \\
Query Generator Dropout &  0.1               & 0.1                  \\
SSP Dropout             &  0.1               & 0.1                  \\
RMLM Dropout            &  0.1               & 0.1                 \\  
FP16 option level       &  O2                & O2                   \\ \bottomrule
\end{tabular}
}
\caption{Hyper-parameters for pre-training.}
\label{tab:hyperparam-pretraining}
\end{table*}

\begin{table*}[]
\centering
\setlength{\tabcolsep}{2.5mm}{
\begin{tabular}{lcccccc}
\toprule
\textbf{HyperParam}     & \textbf{RACE} & \textbf{DREAM} & \textbf{ReClor} & \textbf{MultiRC} &  \textbf{Hotpot QA} \\ \hline
Peak Learning Rate      &  4e-5$^\clubsuit$/2e-5$^\spadesuit$    & 3e-5$^\clubsuit$/2e-5$^\spadesuit$       &  2e-5$^\clubsuit$/1e-5$^\spadesuit$       & 3e-5             & 5e-5$^\clubsuit$/3e-5$^\spadesuit$   \\
Learning Rate Decay     &  Linear       & Linear         &  Linear         & Linear           & Linear                                                       \\
Batch Size              &  32$^\clubsuit$/16$^\spadesuit$        & 24             &  24             & 32               & 32$^\clubsuit$/48$^\spadesuit$                                                   \\
Epoch                   &  4            & 8              &  10             & 8.0              & 3$^\clubsuit$/4$^\spadesuit$                                                          \\
Warmup Proportion       &  0.1$^\clubsuit$/0.06$^\spadesuit$      & 0.1            &  0.1            & 0.1              & 0.1                                                          \\
Weight Decay            &  0.01         & 0.01           &  0.01           & 0.01             & 0.01                                                       \\
Adam $\epsilon$         &  1e-6         & 1e-6           &  1e-6           & 1e-6             & 1e-6$^\clubsuit$/1e-8$^\spadesuit$                                                        \\
Adam $\beta_1$          &  0.9          & 0.9            &  0.9            & 0.9              & 0.9                                                       \\
Adam $\beta_2$          &  0.999$^\clubsuit$/0.98$^\spadesuit$    & 0.999$^\clubsuit$/0.98$^\spadesuit$    &  0.999$^\clubsuit$/0.98$^\spadesuit$      & 0.999            & 0.999     \\
Gradient Clipping       &  1.0$^\clubsuit$/0.0$^\spadesuit$       & 0.0$^\clubsuit$/5.0$^\spadesuit$         &  0.0            & 1.0              & 0.0                                                      \\
Max Sequence Length     &  512          & 512            &  256            & 512              & 384$^\clubsuit$/386$^\spadesuit$                                                         \\
Max Query Length        &  128          & 512            &  256            & 512              & 64                                                        \\ 
Dropout                 &  0.1          & 0.1            &  0.1            & 0.1              & 0.1                                                         \\ \bottomrule
\end{tabular}
}
\caption{Hyper-parameters for fine-tuning. $\clubsuit$: Hyper-parameters for BERT-based models. $\spadesuit$: Hyper-parameters for RoBERTa-based models.}
\label{tab:hyperparam-fine-tuning}
\end{table*}



\section{Analysis of Extra Parameters Introduced}

For fair comparison, we try to introduce as few additional parameters as possible. Since the output layer is highly task-specific and the single head-attention defined in Appendix \ref{sec:simple-attention} is simple, we main analyze the extra parameters introduced for query representation learning defined in \S\ref{query-representation-learning}. 
A single layer of Transformer comprises of a multi-head attention module and a feed-forward network. As a result, the multi-head attention module generating the query representation has introduced 2.8\% extra parameters compared with a 12-layer Transformer without consideration to the parameters in embedding layer and layer normalization. 

\end{document}